\documentclass[conference,12pt]{IEEEtran}
\usepackage{amsfonts}

\usepackage{amsmath,graphicx}
\usepackage{subcaption}
\usepackage{algorithm}
\usepackage{algorithmic}
\usepackage{booktabs}
\usepackage{multirow}
\usepackage{hyperref}
\providecommand{\keywords}[1]{\textbf{\textit{Index terms---}} #1}

\hypersetup{
    pdftitle={Unsupervised Dataset Dictionary Learning for domain shift robust clustering},
    pdfauthor={Anas Hattay, Mayara Ayat, Fred Ngole Mboula},
}

\title{Unsupervised Dataset Dictionary Learning for domain shift robust clustering: application to sitting posture identification}

\author{
    \IEEEauthorblockN{
        Anas Hattay\IEEEauthorrefmark{1},
        Mayara Ayat\IEEEauthorrefmark{2},
        Fred Ngole Mboula\IEEEauthorrefmark{1}
    }
    \IEEEauthorblockA{
        \IEEEauthorrefmark{1}Universit\'e Paris-Saclay, CEA, List, F-91120, Palaiseau, France \\
        Email: anas.hattay@cea.fr, fred-maurice.ngole-mboula@cea.fr
    }
    \IEEEauthorblockA{
        \IEEEauthorrefmark{2}CentraleSup\'elec, Universit\'e Paris-Saclay, F-91190, Gif-sur-Yvette, France \\
        Email: mayara.ayat@student-cs.fr
    }
}

\begin{document}
\maketitle
\pagestyle{plain}

\maketitle
\begin{abstract}
This paper introduces a novel approach, Unsupervised Dataset Dictionary Learning (U-DaDiL), for totally unsupervised robust clustering applied to sitting posture identification. Traditional methods often lack adaptability to diverse datasets and suffer from domain shift issues. U-DaDiL addresses these challenges by aligning distributions from different datasets using Wasserstein barycenter based representation. Experimental evaluations on the Office31 dataset demonstrate significant improvements in cluster alignment accuracy. This work also presents a promising step for addressing domain shift and robust clustering for unsupervised sitting posture identification.
\end{abstract}

\keywords{Unsupervised learning, Domain shift, Robust clustering, Dictionary learning,Optimal transport}

\section{Introduction}
\label{sec:intro}
\subsection{Context}
The digital age has introduced a new era marked by extensive technology use in modern society, contributing to the prevalence of unhealthy posture habits. Prolonged periods of sitting, typically averaging at least five hours a day at a desk, significantly influence posture \cite{article}. Poor posture can result in various issues, including back pain and joint degeneration \cite{tuzun1999low}. To contribute to addressing these concerns, we utilize machine learning to identify sitting postures in real-world settings. 
\subsection{Related Work}
\subsubsection{Sitting Posture Identification}

At present, there are two principal ways of sitting
posture recognition. The initial method employs a camera to capture images of individuals and then recognize their posture (\cite{kulikajevas2021detection} , \cite{9343347}) . An alternative approach involves detecting sitting postures through pressure sensor measurements on the hips, back, or chest (\cite{fan2022deep} , \cite{tsai2023automated}). Both of these paradigms rely on supervised methods, which can be limiting because they do not account for the lack of labels in real-world scenarios. Furthermore, most of these methods do not address the issue of domain shift across different individuals, which is primarily caused by morphological differences. It is important to highlight that these methods commonly train, validate, and test using data from the same individuals, disregarding the wider spectrum of variability across different people. To overcome these limitations, innovative strategies in the following section are worth considering.
\subsubsection{Unsupervised Domain-Adaptive Clustering}

Various methods have been proposed to address the challenge of clustering under domain shift. One prominent approach was introduced by Caron et al. in 2018 \cite{caron2018deep}. It involves training a neural network to extract features from unlabeled data, clustering these features using algorithms like k-means, and then using the resulting cluster assignments as pseudo-labels to refine the neural network parameters iteratively. Currently, the only method that successfully addressed the challenge of clustering under domain shift in real-world scenarios without relying on any labels from the target domain is the one presented by Amit et al \cite{rozner2023domain}. To achieve that, the proposed method follows a two-stage training framework: firstly, employing self-supervised pre-training to extract meaningful features from the data without explicit labels, and secondly, implementing a multi-head cluster prediction approach to assign pseudo labels to the data points based on the extracted features.

In line with this method, we explore the use of an optimal transport-based method to cluster data across various domains in a completely unsupervised manner. Such an approach is very useful when performing multi-domain clustering without resorting to data from target domain, a common scenario in real-life applications often overlooked in previous methods.

\section{Our Work}
\label{sec:method}

\subsection{Background}

\subsubsection{Optimal Transport}

The principal idea of optimal transport lies in finding the optimal mapping between source and target domains that minimizes the overall cost or distance while satisfying the conservation of mass constraint.

\textbf{Mathematical Formulation}
Let \(x^{(P)}_i \overset{\text{i.i.d.}}{\sim} P\) (resp. \(x^{(Q)}_j \overset{\text{i.i.d.}}{\sim} Q\)). We can empirically approximate \(P\) and \(Q\) using mixtures of Dirac deltas,
\begin{equation}
\hat{P}(x) = \frac{1}{n_P} \sum_{i=1}^{n_P} \delta(x - x^{(P)}_i).
\label{eq:myequation}
\end{equation}
\begin{equation}
\hat{Q}(x) = \frac{1}{n_P} \sum_{j=1}^{n_P} \delta(x - x^{(Q)}_j).
\label{eq:myequation}
\end{equation}

Hereafter, we refer to \(\hat{P}\) as the point cloud. We denote \(X^{(P)} = [x^{(P)}_1, \ldots, x^{(P)}_{n_P}] \in \mathbb{R}^{n_P \times d}\) as the support of \(\hat{P}\). The Kantorovich problem aims to find an OT plan, \(\pi \in \mathbb{R}^{n_P \times n_Q}\), where \(\pi_{i,j}\) indicates how much mass \(x^{(P)}_i\) should send to \(x^{(Q)}_j\) is such a way to minimise the overall transportation cost. The plan \(\pi\) must preserve mass, meaning,

\begin{equation}
\Pi(\hat{P}, \hat{Q}) := \left\{ \pi : \sum_{i} \pi_{i,j} = \frac{1}{n_Q}; \sum_{j} \pi_{i,j} = \frac{1}{n_P} \right\}
\end{equation}

In this context, the OT problem between \(\hat{P}\) and \(\hat{Q}\) can be formulated as,

\begin{equation}
\pi^\star = \text{OT}(X^{(P)}, X^{(Q)}) = \text{argmin}_{\pi \in \Pi(P^\wedge, Q^\wedge)} \langle C, \pi \rangle_F
\end{equation}

where \(\langle\cdot, \cdot\rangle_F\) denotes the Frobenius inner product and \(C_{i,j} = c(x^{(P)}_i, x^{(Q)}_j)\) is referred to as the ground cost matrix.

\subsubsection{Wasserstein Barycenter}

The Wasserstein barycenter is a Frechet mean of a set of distributions for the Wasserstein distance, which is a distribution that minimizes the weighted sum of Wasserstein distances from the set of input distributions. 

\textbf{Mathematical Formulation}
For distributions \(P = \{P_k\}_{k=1}^K\) and weights \(\alpha \in \Delta_K\), with \(\Delta_n = \{a \in \mathbb{R}_+^n : \sum_i a_i = 1\}\) , and \(B(\cdot; P)\) represents the barycentric operator, the Wasserstein barycenter is a solution to :
\begin{equation}
B^{\star} = B(\alpha; P) = \inf_B \sum_{k=1}^K \alpha_k W_c(P_k, B).
\label{equation:five}
\end{equation}

\subsubsection{Dataset Dictionary Learning (DaDiL)}

 DaDiL \cite{inbook} is a dictionary learning technique that uses optimal transport to represent input distributions as Wasserstein barycenters of learnt probability distributions referred to as "atoms." This involves training a dictionary with these atoms and minimizing the weighted sum of the reconstruction error to align with the training distributions.

\subsection{Approach}

To tackle the challenge of jointly finding clusters in domain shifted datasets, we propose a novel approach, illustrated in figure \ref{fig:Training}.
\subsubsection{Training using source domains data }

\begin{enumerate}
    \item \textbf{\textit{Generating Pseudo-Labels:
     }}To set the stage for DaDiL, which requires labeled input data, we utilized K-Means with random initialization to generate pseudo labels.\item \textbf{\textit{Cluster Alignment: 
}}
Next, an additional step is required to rearrange the pseudo-labels due to the absence of knowledge regarding cluster correspondence across different domains. To accomplish this, a cost matrix is computed to quantify the transportation cost between pairs of clusters, utilizing the Wasserstein distance to quantify cluster dissimilarities.

%The transportation cost from one cluster in a domain to another is mathematically represented and a cost matrix is obtained.

%The cluster matching problem is formulated as a linear optimization task to minimize the total transportation cost.

%The objective is to find optimal correspondences between clusters in one domain and clusters in another domain that minimize the total transportation cost.

The optimal matching between two clusters is found solving the following problem:

%\textit{Objective Function:} 

\begin{equation*}
\underset{x_{ij}}{\min} \sum_{i=1}^{n} \sum_{j=1}^{n} c_{ij}x_{ij}
\end{equation*}

subject to the following constraints: 
\begin{align*}
&\sum_{j=1}^{n} x_{ij} = 1, \text{ for } i = 1, 2, \ldots, n, \\
&\sum_{i=1}^{n} x_{ij} = 1, \text{ for } j = 1, 2, \ldots, n.
\end{align*}

%where $c_{ij}$ represents the cost associated with mapping source cluster $i$ in domain 1 to target cluster $j$ in domain 2, subject to the assignments of source cluster $k$ in domain 1 to target cluster $l$ in domain 2. 
where $c_{ij}$ is the Wasserstein distance between the cluster $i$ in domain 1 and the cluster $j$ in domain 2, $n$ is the number of clusters in the two domains and the $x_{ij}$s are the \textit{Decision Variables:}

\begin{equation*}
x_{ij} =
\begin{cases}
1, & \text{if source cluster } i \text{ in domain 1 is } \\
& \text{mapped to target cluster } j \text{ in domain 2} \\
0, & \text{otherwise}.
\end{cases}
\end{equation*}.
%for $i, j = 1, 2, \ldots, n$, representing the mapping decisions.

%\textit{Constraints:}
%\begin{align*}
%&\text{Each source cluster in domain 1 is mapped to exactly } \\ & \text{one target cluster in domain 2:} \\
%&\sum_{j=1}^{n} x_{ij} = 1, \text{ for } i = 1, 2, \ldots, n, \\
%&\text{Each target cluster in domain 2 is mapped to exactly} \\ & \text{one source cluster in domain 1:} \\
%&\sum_{i=1}^{n} x_{ij} = 1, \text{ for } j = 1, 2, \ldots, n.
%\end{align*}

The cluster alignment enables the sharing of pseudo-labels between the two domains. This process is repeated for all domains within the dataset and one domain is taken as a reference.

\item \textbf{\textit{Atoms Initialization:}} The number of atoms is set to match the number of domains. They are initialized as slightly shifted versions of each domain dataset towards the others.

%To initialize dictionary atoms, we compute a Wasserstein barycenter across domains, obtaining weight coefficients that quantify the contribution of each domain. Using these weights, we create individual barycenters for each domain, which serve as initialized atoms for training our dictionary.

\item \textbf{\textit{Dictionary Training:}} We proceed to train a dictionary using K-Means derived pseudo labels, with atoms initialized as described earlier. At the end of this step, we obtain optimized atoms and optimized coefficients that describe the relative contribution of each of these atoms in the reconstruction of each domain’s data. 

%Numerous hyperparameters may be fine-tuned to achieve optimal results.

\item \textbf{\textit{Centroids Calculation:}} Using the current dictionary, the clusters are refined as follows: each source domain is approximated using a Wasserstein barycenter of the atoms using the domain's barycentric coordinates obtained from dictionary training. The number of points in each barycenter's support is set to the required number of clusters $n$ so that each point is a candidate prototype member of a cluster in the domain.
These prototypes are further displaced using barycentric mapping~\cite{flamary2016optimal}, in such a way to improve their coverage of the considered domain. The resulting synthetic data points are used as centroids to compute new clusters for each source domain.

%At this stage, we want to perform clustering on the training domain data by exploiting the resulting atoms. This will later help us improve the pseudo-labels and further refine the atoms based on the results of this clustering. To achieve this, 

%Instead of selecting a number of points in the support of the barycenter equal to the domain features dimension, which would offer us a reconstruction similar to the target domain features, we set the number of points in the barycenter's support to the desired number of clusters. Consequently, we obtain new points that could be taken as prototypes for each cluster. We directly move them to ensure better significance. To achieve this, we update the centroid by averaging the points it has sent mass to, weighted by the amount of mass sent.

\item \textbf{\textit{Cluster Assignment: }} After this step, similar to how k-means operates, we calculate the Euclidean distance between
each point and each centroid for each domain data. We then assign each point to the cluster
represented by the nearest centroid.
\item \textbf{\textit{Iterative Process:}}
To establish correspondences between similar classes in different domains, we use a second cluster alignment process. Then, we reiterate the procedure using updated pseudo-labels.

\begin{figure}
    \centering
    \resizebox{\columnwidth}{!}{%
        \includegraphics{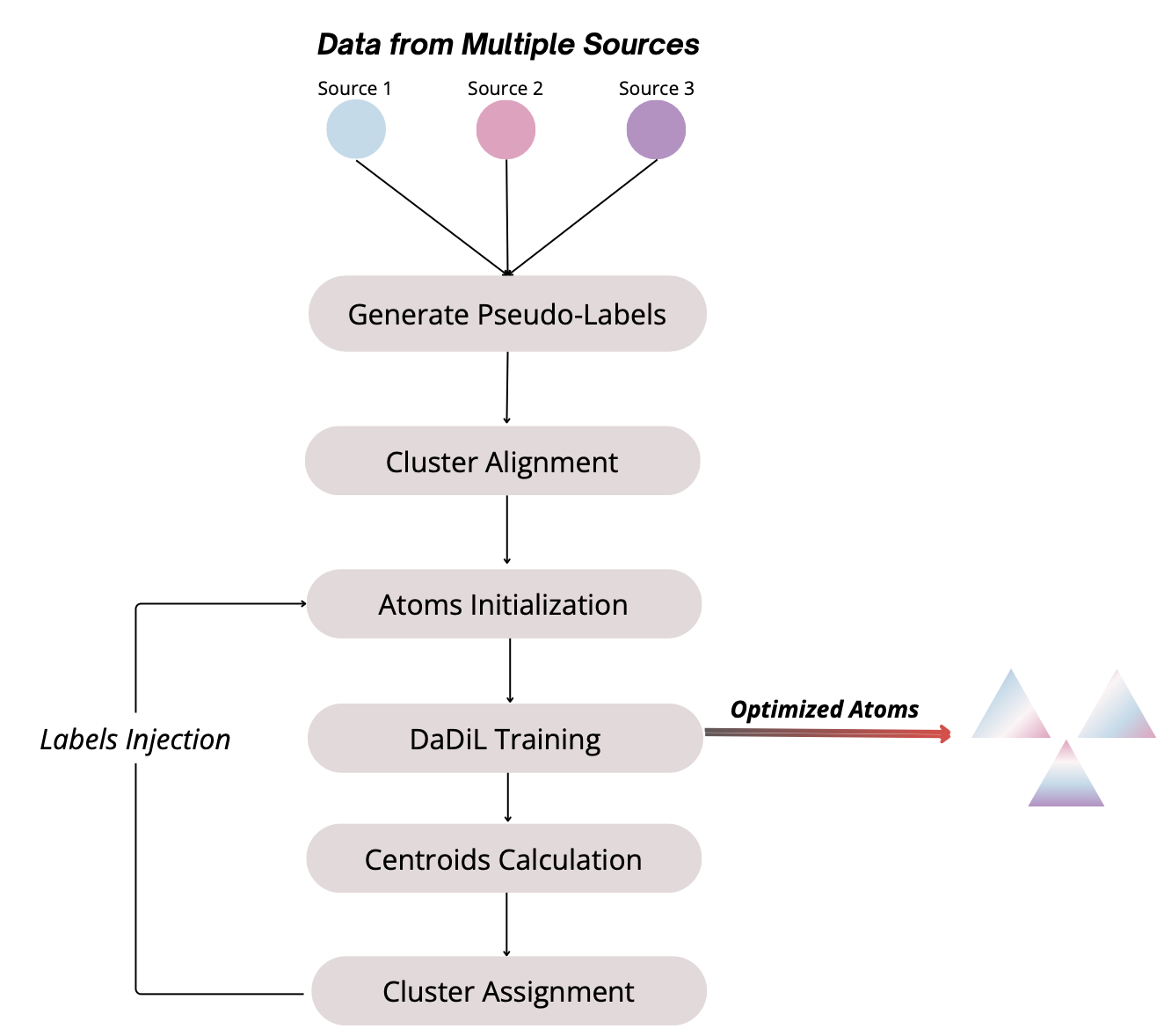}
    }
    \caption{Training pipeline on source domains.}
    \label{fig:Training}
\end{figure}

\subsubsection{Inference using target domain data:}
In our approach, the inference process relies on exploiting the optimized atoms obtained during the training process to ultimately derive centroids for clustering. As illustrated in Figure \ref{fig:target}, the adaptation occurs through performing a barycentric regression between these atoms and the features of the target domain. As a result of selecting a number of points in the support of the barycenter equal to the desired number of clusters, we obtain new points, which are the target domain clusters prototypes. Similar to the fifth step of the training, these prototypes are refined and used as centroids to construct clusters.

%Ultimately, we will have our centroids that can be used to assign features from the target domain to different clusters.

%This data is preprocessed in the same way as the training domains which means we employ the same feature-extraction method.

%Then, a barycentric regression between the optimized atoms and the target features results in the determination of a new barycenter. This process yields clusters along with centroids with respect to the fixed number of clusters.
\begin{figure}
    \centering
    \resizebox{\columnwidth}{!}{%
        \includegraphics{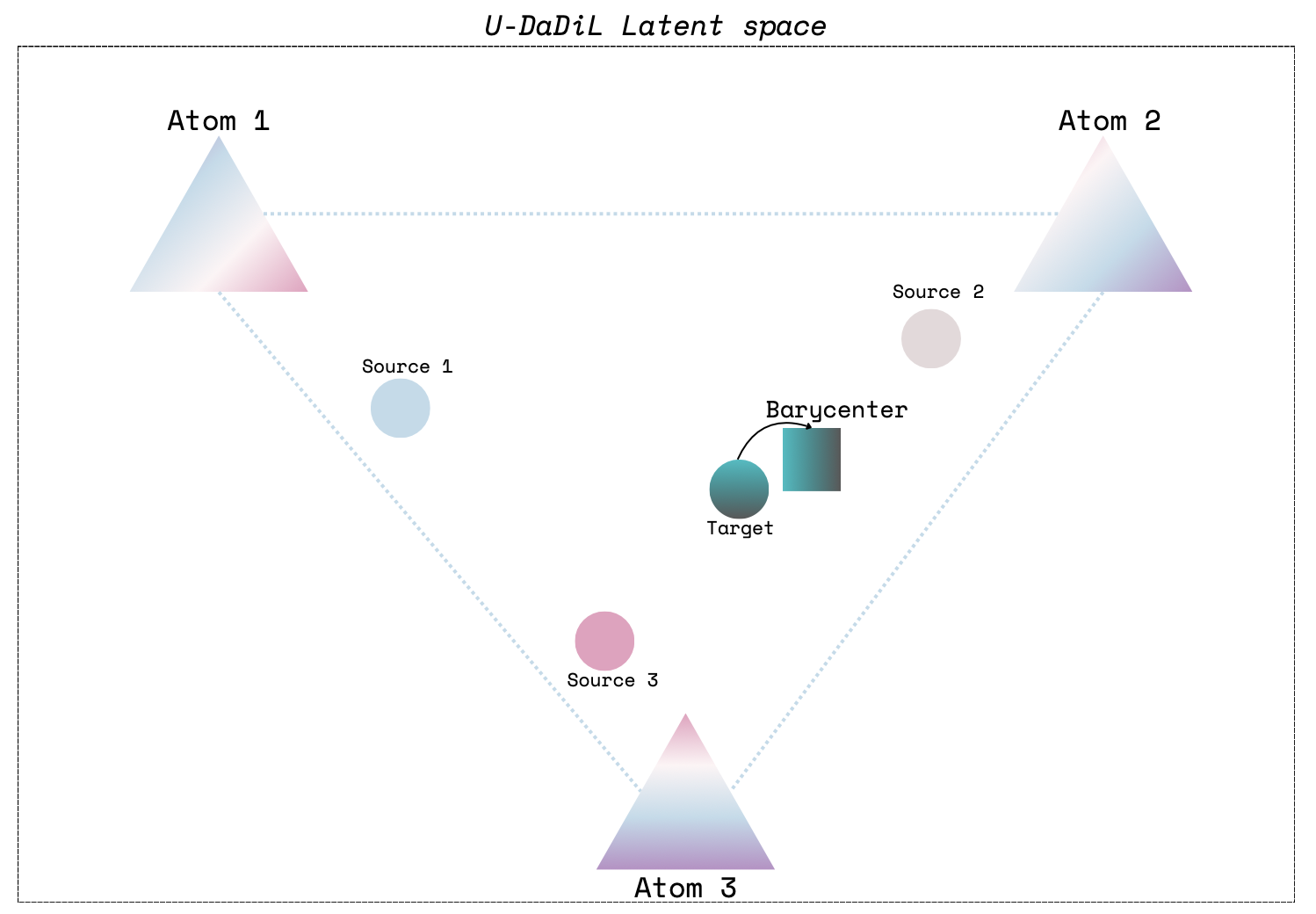}
    }
    \caption{Adaptation process on target domain.}
    \label{fig:target}
\end{figure}

\end{enumerate}

\section{Numerical Experiments}
\label{sec:numexp}

\subsection{Benchmark Dataset - Office31 :}

We evaluated our framework on the widely used Office31 dataset \cite{10.1007/978-3-642-15561-1_16}, which consists of 31 categories of office objects distributed across three domains: Amazon, DSLR, and Webcam. 
We extracted features using ResNet50 pre-trained model and then compared our method with ACIDS \cite{menapace2020learning}, Invariant Information Clustering for Unsupervised Image Classification and Segmentation \cite{ji2019invariant},  DeepCluster \cite{caron2018deep} , Domain-generalizable multiple-domain clustering \cite{rozner2023domain} and K-Means \cite{lloyd1982least}. Experimental setting was the same as in~\cite{rozner2023domain} in order to make U-DaDiL results comparable to those established in that work.

\subsection{Sitting Posture Identification}
In order to evaluate the method in posture identification, we collected pressure map data on seven distinct known seating positions across five subjects, with different weights and heights. 
Therefore each subject data are considered to represent a single domain. %The experiments utilized data gathered from five subjects,  
We trained U-DaDiL on these domains and compared the clustering results with those of the K-Means Baseline.

%and performed clustering using the resultant centroids. Given the interesting outcomes from K-Means, we 

%\begin{table}[h!]
 %   \centering
    
  %  \resizebox{1\columnwidth}{!}{%
   % \begin{tabular}{lccc}
    %    \toprule
     %   \textbf{Subject} & \textbf{Weight (kg)} & \textbf{Gender} & \textbf{Height (cm)} \\
       % \midrule
        %Subject 1 & 57 & Female & 165 \\
        %Subject 2 & 74 & Female & 164 \\
        %Subject 3 & 88 & Male & 175 \\
        %Subject 4 & 80 & Male & 200 \\
       % Subject 5 & 95 & Male & 175 \\
        %\bottomrule
    %\end{tabular}%
    %}
    %\caption{Summary of Subjects}
    %\label{tab:subjects}
%\end{tab

% We handle the empty clusters, that occur in the mapping between the found centroids and the data points, by diving the largest cluster in the domain. We randomly alter its centroid to obtain the new centroid. We then reassign the data points in the cluster to the two resulting clusters. \cite{caron2018deep}


\section{Results and Discussions}

\subsection{Results on Office-31}

In Table \ref{tab:method_comparison}, mirroring the approach of the recent method by Amit et al., we assess our method's performance per target class against deep clustering state-of-the-art methods on the Office-31 dataset. We conducted tests with three-domain combinations, training the method each time on two source domains and using the third as the target domain, whose features are entirely unseen during the training phase. As shown in Table~\ref{tab:method_comparison}, the proposed method surpasses all previous approaches across all three domains combinations, demonstrating an average improvement of 24.6\% compared to the best results reported in the literature.

\begin{table}[htbp]
    \centering
    \resizebox{1\columnwidth}{!}{%
    \begin{tabular}{lcccccc}
        \toprule
        Method & Target Fine-tuned & Supervision &  D, W $\rightarrow$ A & A, W $\rightarrow$ D & A, D $\rightarrow$ W & Avg\\
        \midrule
        DeepCluster (Caron et al., 2018) & $\surd$ & - & 19.6 & 18.7 & 18.9 & 19.1 \\
        IIC (Ji et al., 2018) & $\surd$ & - & 31.9 & 34.0 & 37.0 & 34.3 \\
        IIC-Merge (Ji et al., 2018) & $\surd$ & - & 29.1 & 36.1 & 33.5 & 32.9 \\
        IIC + DIAL (Ji et al., 2018) & $\surd$ & - & 28.1 & 35.3 & 30.9 & 31.4 \\
        Continuous DA (Mancini et al., 2019) & $\surd$ & - & 20.5 & 28.8 & 30.6 & 26.6 \\
        ACIDS (Menapace et al., 2020) & $\surd$ & - & 33.4 & 36.1 & 37.5 & 35.6 \\
        K-means (Lloyd, 1982) & $\surd$ & - & 14.9 & 24.3 & 20.8 & 29.9 \\
        DGMDC (Amit et al., 2024) & - & - & 24.1 & 50.1 & 47.7 & 40.6 \\
        \textbf{U-DaDiL (Ours)}  & - & - & 34.4 & 64.4 & 62.8 & 50.6 \\
        \bottomrule
    \end{tabular}%
    }
    \caption{Comparison of state-of-the-art methods based on clustering accuracy results for the Office31 dataset across all three domain combinations: Amazon (A), Webcam (W), and DSLR (D). The notation X,Y → Z denotes training on domains X and Y and testing on domain Z. "Target fine-tuned" denotes that data from the test domain was incorporated during either the training or an adaptation process.} 
    \label{tab:method_comparison}
\end{table}

% To visualize U-DaDiL response , we utilized the t-SNE algorithm on both the features extracted from Amazon domain data and the centroids obtained through reconstruction after the training phase. The results are depicted in Figure \ref{fig:Centroids}.

% \begin{figure}
%     \centering
%     \resizebox{\columnwidth}{!}{%
%         \includegraphics{images/centroidesoverdata.png}
%     }
%     \caption{t-SNE of U-DaDiL response for Webcam domain (red: Centroids, blue: Webcam Resnet50 Features).}
%     \label{fig:Centroids}
% \end{figure}

\subsection{Results on Posture Data}
Table \ref{tab:3subjects}  
presents the performance of U-DaDiL on different subjects i.e domains. Across these domains, our method consistently  showcased strong performance compared to K-Means, which occasionally produced less optimal clustering results.

\begin{table}[htbp]
    \centering
    \resizebox{1\columnwidth}{!}{%
        \begin{tabular}{lcccccc}
            \toprule
            & Domain 1 & Domain 2 & Domain 3 & Domain 4 & Domain 5\\ 
            \midrule
            K-Means & 1.0 & 1.0 & 1.0 & 0.836 & 0.769 \\ 
            U-DaDiL & 0.953 & 1.0 & 1.0 & 1.0 & 0.909 \\ 
            \bottomrule
        \end{tabular}%
    }
    \caption{Comparison of ARI results using U-DaDiL and K-Means on different subjects/domains.}
    \label{tab:3subjects}
\end{table}

% \begin{figure}[htbp]
%   \centering
%   \begin{subfigure}[b]{0.45\textwidth}
%     \centering
%     \includegraphics[width=\textwidth]{dadil_data1.png}
%     \caption{t-SNE Visualization for the first domain using U-DaDiL}
%     \label{fig:sub1}
%   \end{subfigure}
%   \hfill
%   \begin{subfigure}[b]{0.45\textwidth}
%     \centering
%     \includegraphics[width=\textwidth]{kmeans_datacentroids_1.png}
%     \caption{t-SNE Visualization for the first domain using K-Means}
%     \label{fig:sub2}
%   \end{subfigure}
 
%   \label{fig:general}
% \end{figure}

\subsection{Discussion}

We believe that this improvement is due to our method underlying regularization: by learning shared atoms across domains, U-DaDiL implicitly takes advantage of domain similarities to identify centroids that are more representative of the true organisation of the data.

%U-DaDiL harmonizes between distribution and geometry leading to centroids that are more representative of the true nature of the data.

In fact, to do clustering, we do not only rely on spatial proximity but also consider the distributional properties of data points since our optimization process aims to minimize the Wasserstein distance between the reconstructed distribution and the original data distribution, taking into consideration the inclusion of multi-domain information. This dimension ensures that our centroids adapt not only to the shape, spread and other statistical properties of the data within individual domains but also to the complex interplay between these domains.

%\vfill
%\pagebreak

\section{Conclusion}
\label{sec:ref}

This work was centered on addressing the challenge of characterizing sitting postures in real-world settings by developing novel unsupervised cross-domain clustering method.
To illustrate the effectiveness of our methodology, we conducted a case study using the Office31 dataset. Our clustering results outperformed other clustering methods, demonstrating a remarkable average of 24.63\% improvement in accuracy compared to the best method results. Additionally, we applied our method to a real-world application using pressure map posture data and observed better posture clustering results compared to the K-Means baseline.

Future research directions include tackling heterogeneous multi-domain clustering with varying clusters numbers across domains.

%In future research, we anticipate that unbalanced optimal transport could offer solutions in specific scenarios, enabling multi-domain clustering even in cases with varying cluster numbers across domains.

\bibliographystyle{IEEEbib}
\bibliography{refs}
\end{document}